\documentclass[letterpaper,twocolumn,fleqn]{article}

\usepackage{ist}
\usepackage{times}
\usepackage{epsfig}
\usepackage{graphicx}
\usepackage{amsmath}
\usepackage{amssymb}
\usepackage{comment}
\usepackage{multirow}
\usepackage{biblatex}

\title{Efficient Temporally-Aware DeepFake Detection using H.264 Motion Vectors}

\author{Peter Grönquist*, Yufan Ren*, Qingyi He, Alessio Verardo, Sabine Süsstrunk \\
Image and Visual Represenation Lab, École Polytechnique Fédérale de Lausanne; Lausanne, Switzerland \\
* Equal Contribution}

\date{} 

\bibliography{egbib}

\begin{document} 

\maketitle 

\thispagestyle{empty} 


\begin{abstract}
Video DeepFakes are fake media created with Deep Learning (DL) that manipulate a person's expression or identity. Most current DeepFake detection methods analyze each frame independently, ignoring inconsistencies and unnatural movements between frames. Some newer methods employ optical flow models to capture this temporal aspect, but they are computationally expensive. In contrast, we propose using the related but often ignored Motion Vectors (MVs) and Information Masks (IMs) from the H.264 video codec, to detect temporal inconsistencies in DeepFakes. Our experiments show that this approach is effective and has minimal computational costs, compared with per-frame RGB-only methods. This could lead to new, real-time temporally-aware DeepFake detection methods for video calls and streaming.
\end{abstract}

\section{Introduction}
\label{sec:intro}



Ever since their emergence, DeepFakes, or more specifically Deep Learning (DL)-based fake media with manipulated facial identity and expression, have brought serious security and privacy concerns to the public~\cite{originaldeepfakes,verdoliva2020media}. 
Whilst they offer advantages to photo and video editing, they also have significant downsides, including identity abuse and the spreading of misinformation. 
Ranging from defamation and political misuse to video-call scams and pornography, these issues have become a major cause for concern in our society~\cite{DeepfakeScam}; even more so as facial identity is a determining factor for person recognition, both in the human thought process and in machine algorithms.~\cite{galbally2014biometric}. 
Given their importance in our social interactions~\cite{frith2009role}, a simple modification of facial expressions can have profound effects in how a person is perceived.


\begin{figure}[!t]
\setlength{\belowcaptionskip}{-0.375cm}
    \centering
    \includegraphics[width=0.5\textwidth,page=7]{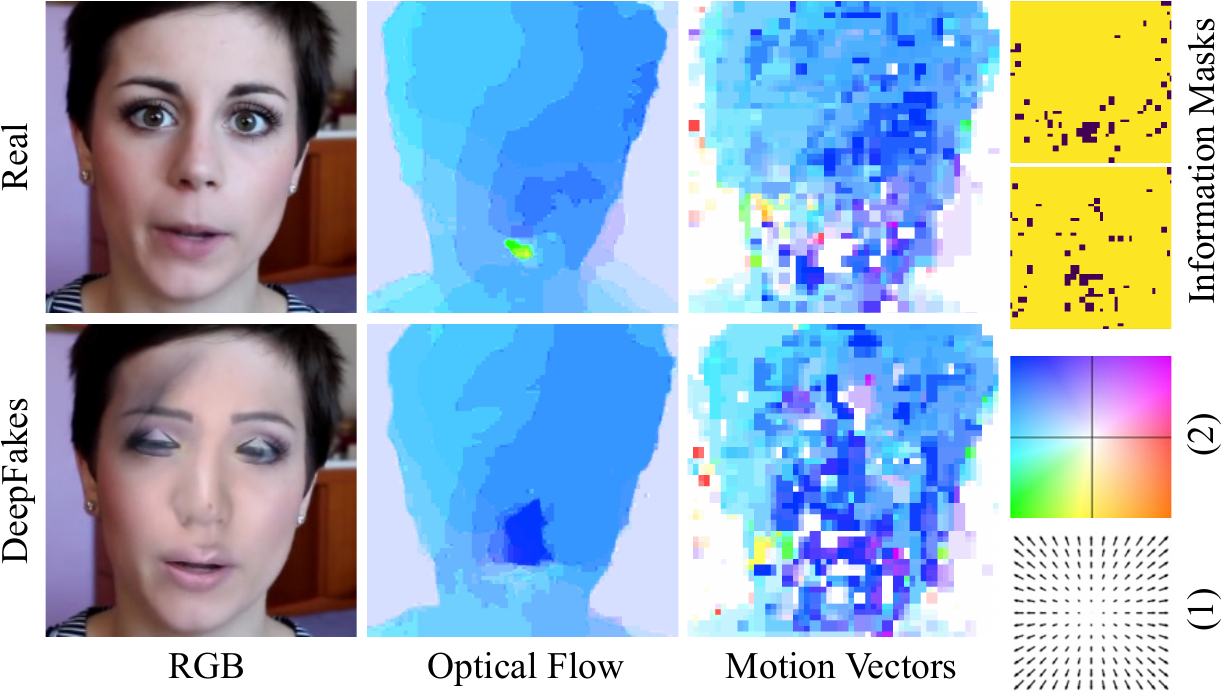}

    \caption{Two continuous video frames in the FaceForensics++ dataset, with their corresponding optical flow and H.264 motion vectors, with respect to the precursory frame of each frame. The information masks indicate the availability of motion vectors at a spatial location. (1) and (2) express the color visualization of two dimensional motion information into RGB space. Motion vectors show similar motion as optical flow but are coarser and noisier. (Best viewed on a screen when zoomed in)}
    
\label{fig:teaser}
\vspace{-15pt}
\end{figure}

Therefore, to combat DeepFake videos, various research groups, companies and organizations have launched multiple campaigns to raise public awareness of these issues~\cite{TrustedMedia,kaggleDFDC}. These initiatives have notably accelerated research on DeepFake countermeasures, demonstrating significant improvements in accuracy, efficiency, and generalizability~\cite{cozzolino2017recasting,rossler2018faceforensics, rossler2019faceforensics++,li2020face, shiohara2022detecting, zhang2022deepfake,qi2020deeprhythm,ciftci2020fakecatcher, ciftci2020hearts,agarwal2020detecting,amerini2019deepfake, caldelli2021optical}. 
%
%
%
However, despite these achievements, most contemporary DeepFake detection methods exhibit two key limitations: the omission of temporal video information and insufficient generalizability~\cite{zhang2021detecting}.

With the fast-paced evolution of DeepFake generation models, it is necessary for DeepFake detection algorithms to be more robust to these unknown manipulations, as measured by cross-forgery accuracy. 
Current such attempts at improving generalizability focus mostly on finding new fake cues that are as algorithm-independent as possible, such as detecting heartbeat rhythms~\cite{qi2020deeprhythm, ciftci2020fakecatcher, ciftci2020hearts} or blending boundaries~\cite{li2020face}. 
However, these come with their own restrictions of needing a specific quality of videos or not having the video fully generated.

Handling temporal information poses its own specific challenges; a typical DeepFake detection pipeline will sample multiple frames from a video, predict per-frame fake probabilities, and then heuristically aggregate these probabilities into an overall fake video probability. 
This method, however, fails to account for the inherent temporal consistency stemming from real-world constraints, such as stable facial features, unchanged eye colors, and naturally paced blinking. One common way of capturing this temporal information is to use the motion information in videos, commonly represented as Optical Flow (OF)~\cite{caldelli2021optical}. However, one issue is that optical flow estimation requires additional computational resources, in a sequential manner, which poses a potential efficiency bottleneck. This is especially the case considering the increasingly growing prevalence of DeepFakes in live streaming and online video platforms.

Based on the two above-mentioned limitations, this paper proposes the use of H.264 motion vectors as a method for motion approximation in DeepFake detection: On one hand, H.264 motion vectors offer new and unique fake artifacts and MV-based models are shown to be more generalizable than per-frame RGB-only methods. On the other hand, we do not need to estimate motion vectors as in optical flow-based methods, which is a benefit provided by the widespread adoption of the H.264 codec. 
%
%
In research on using MV for segmentation~\cite{SOLANACIPRES200999}, the motion approximation capabilities of MVs and OF are compared. They suggest that MVs can provide a good estimate for optical flow, except for small noise, which can for example stem from imperceptible variations in luminosity. 
In those cases, MVs might reference different patches, similar in luminosity and color, that are contrary to the optical flow, or generally just plain noisy. 
It is exactly these temporal inconsistencies, we hypothesize, that might present a challenge for generative DeepFake methods, and which we can exploit as an auxiliary input for detection.

To keep model performance high on top of keeping computational costs low, we propose a classification model based on MobileNet~\cite{howard2017mobilenets,howard2019searching}. 
%
To demonstrate the effectiveness of our framework, we conduct rigorous experiments on the FaceForensics++ dataset~\cite{rossler2019faceforensics++}, which consists of several different types of DeepFake generated content. 
In these experiments, and compared with our optical flow baseline classifier, our motion vector based model achieves a relative improvement of $\sim$14\% in accuracy. 
We summarize our findings with the following contributions:

\begin{itemize}
    \item We propose a novel DeepFake detection framework that takes into account temporal transformations and artifacts, outperforming state-of-the-art optical flow based models.
    \item We reach real-time data extraction efficiencies by using readily available H.264 motion vectors as motion approximation.
    \item We demonstrate that our proposed method achieves higher generalization capabilities than models based solely on RGB input, and rival combined RGB and optical flow based models.
\end{itemize}

\section{Related work}
\label{related_work}

\begin{figure*}[!ht]
    \centering
    \includegraphics[width=1.0\textwidth, page=3]{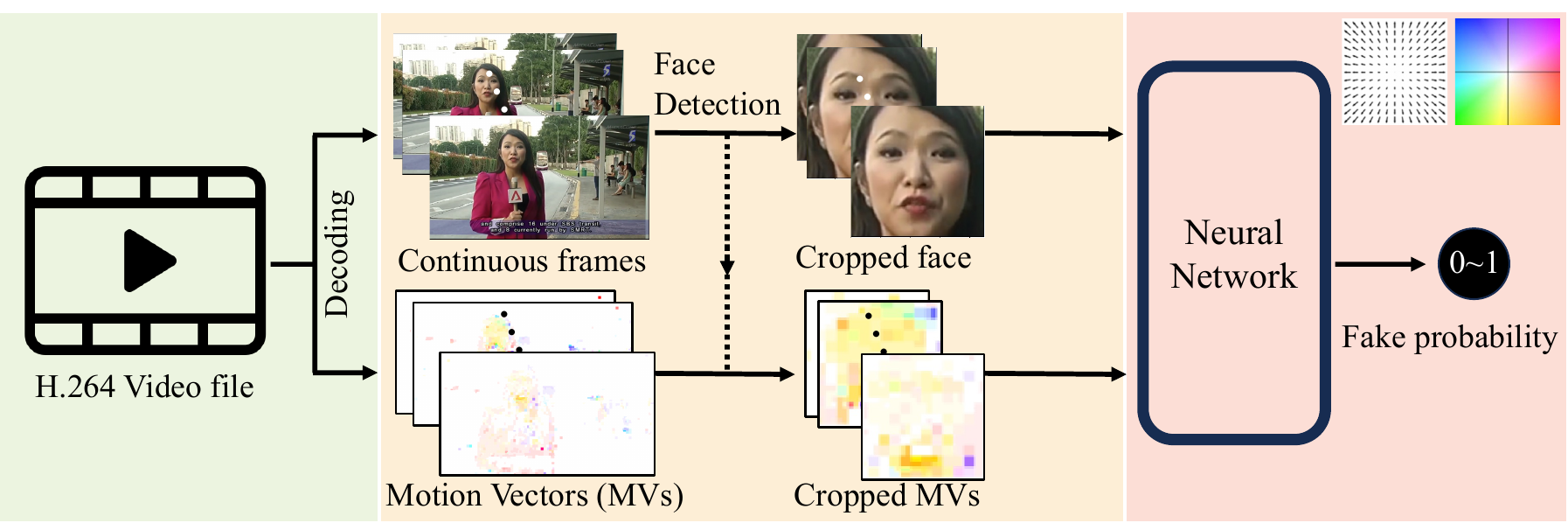}
    
    \caption{
    \textbf{Proposed DeepFake detection pipeline}. The input to the processing set-up is an H.264 encoded video-file or stream. The file or stream are then further processed by selecting a subset of the RGB-frames and cropping the face via a face-detection network. During the decoding process, the motion vectors and a mask of the I-macroblocks are taken directly from the decoder, cropped to the bounding box and used as input to either the standalone MobileNet, or our Two-Stream network.
    The final inputs to the networks vary depending on the transformations performed on the motion vectors and the network used. For example we can use temporal slices of varying duration, here multiple frames, alone, with RGB, and with the I-macroblock mask. Information masks are omitted for simplicity. (Best viewed on a screen when zoomed in)
    }

\label{fig:pipeline}
\vspace{-15pt}
\end{figure*}

\textbf{DeepFakes Detection.} Based on whether they use the inter-frame information, we categorize DeepFakes detection algorithms into two types: image-level and temporally-aware. 
Image-level detections exploit intra-frame information, such as forensics features \cite{cozzolino2017recasting}, XceptionNet features \cite{rossler2018faceforensics, rossler2019faceforensics++}, and blending boundary~\cite{li2020face, shiohara2022detecting}. 
However, image-level algorithms fail to use temporal cues in videos, such as cross-frame inconsistency \cite{zhang2022deepfake}, heartbeat rhythms \cite{qi2020deeprhythm, ciftci2020fakecatcher, ciftci2020hearts}, phoneme-viseme mismatching \cite{agarwal2020detecting}, and unnatural movement \cite{amerini2019deepfake, caldelli2021optical}. 

\textbf{Temporally-aware DeepFake Detection.} To exploit temporal information, researchers proposed and adapted deep networks that can process multiple frames from the computer vision community, such as 3DCNNs \cite{zhang2021detecting}, spatiotemporal transformers \cite{zhang2022deepfake}, 
recurrent networks \cite{guera2018deepfake}, and LSTMs \cite{amerini2020exploiting}. 
Comparatively, instead of feeding all frames as concatenated input to the same network, research by Simonyan and Zisserman claims there is an advantage in using two-stream networks. These networks that have two branches for encoding inputs, feed in pre-extracted motion information, i.e.\ optical flow, in a separate branch from the RGB input, and have for example been found to help with action classification~\cite{simonyan2014two}.
Optical flows, in this case, describe object motion, and we consider it as one of the most general forms of motion representation for DeepFake detection \cite{amerini2019deepfake, caldelli2021optical}. 

One issue that might however hinder the use of optical flow for DeepFake detection is the computational cost of optical flow estimation. 
Despite the efforts that have been made to speed up optical flow estimation to near real-time or real-time\footnote{\url{https://developer.nvidia.com/opticalflow-sdk}}~\cite{DBLP:journals/corr/KroegerTDG16,teed2020raft}, the computational costs in DeepFake detection are still quite large, as OF has to be calculated for every single frame of a video. Therefore it is beneficial to further reduce the resource usage. 
In this paper, instead of trying to speed up optical flow estimation, we use the almost free motion vectors in the H.264 video codec as an approximation to the optical flow for DeepFakes detection.

\textbf{Motion vectors as motion approximation.} H.264 uses motion vectors to exploit temporal redundancy in video frames for compression purposes. 
Motion vectors bear similarity to the optical flow, in that they are both two-dimensional data describing pixel or patch level motion across frames. This similarity and the motion approximation relation has been used for a long time, for example in video object detection, researchers use motion vector to propagate object detection result of key frames~\cite{yokoyama2009motion}. 
Motion vectors have also been used to represent a compressed version of videos, that can be used as input directly~\cite{wu2018compressed}, reducing the data size by up to two orders of magnitude.
In our case, we use the motion itself as a discriminative feature for DeepFake detection, by similarly making use of the compressed temporal information. 
%




\section{Preliminaries and Methodology}
\label{method}

In this Section, we first briefly introduce optical flow and motion vectors. After that, we discuss our data processing methods, as well as the classifier. The whole detection pipeline is illustrated in Fig.~\ref{fig:pipeline}. 

\subsection{Optical flow}
\label{method-opticalflow}
Optical flows represent the projection of every 3D point's trace to the image plane.
The motion information they provide is very helpful for other computer vision tasks, such as detecting and tracking objects~\cite{shin2005optical}, and visual odometry~\cite{muller2017flowdometry}.
However, extracting optical flow from a video stream is one of computer vision's unsolved tasks, due to its ill-posed nature. Examples include having to deal with occlusion of objects, non-rigid movement or even blurry or noisy images.

Over time there have been several proposed solutions to optical flow estimation, for example, nowadays classical optical flow estimations algorithms solve the ill-posed aperture problem by introducing a smoothness prior, such as the wildly used optical flow estimation algorithm TV-L1~\cite{perez2013tv, zach2007duality}.
Researchers also employ more recent deep learning technology to exploit learnable and generalizable priors, achieving state-of-the-art performance on multiple optical flow estimation datasets such as Sintel~\cite{Butler:ECCV:2012} and KITTI 2015~\cite{Menze2018JPRS}.
Therefore, for a fair comparison with our methods, we use the RAFT model which has a top score on both of these datasets, as our optical flow baseline~\cite{teed2020raft}.

\subsection{H.264 motion vectors}
\label{method-mvs}



Motion Vectors (MVs) are part of the H.264\footnote{\url{https://www.itu.int/rec/T-REC-H.264-202108-I/en}} video compression scheme that exploits temporal and spatial redundancy for a better compression rate. 
Unlike optical flow which gives a movement prediction for each pixel, motion vectors operate on macro-blocks, consisting of 4x4, 8x8 and 16x16 pixel blocks, allowing for mixed sizes such as 4x8 and 16x8. 

To fully understand the computational advantage of using encoded MVs, we shortly recap their encoding and decoding process.
There exist three types of frames in the H.264 format: Intra-coded (I), Predicted (P), Bi-directionally predicted (B). 
Of the three, the I-frames do not interest us, as they contain no temporal information and are fully intra-coded.
The P- and B-frames both reference other frames for their encoding and decoding, with the difference being that P-frames reference only past frames and B-frames reference past and future frames. Both are however limited to a maximal distance of 16 frames from the current frame. Therefore, we can gain temporal information, here MVs, from P- and B-frames. 

In P- and B-frames, MVs are not always available for all macro-blocks. Some macro-blocks are encoded entirely with intra-frame coding, so called I-Macroblocks. We use a binary mask to represent these and call the result our Information Mask (IM), meaning there is no temporal dependency in those macroblocks and all information is \textit{new}. We do not delve deeper into how the I-Macroblocks are coded, as only the encoder's incapability of encoding the information with MVs interests us.

Another interesting process is how this temporal information is encoded and decoded. For this the H.26X compression schemes follow a motion estimation algorithm: Firstly, MVs between the to-be-encoded macroblocks are predicted. This can be done via diverse easy-to-compute mechanisms (e.g. median of previous, in encoding order, neighbours) and vary depending on the H.26X version. Then, the actual MV \textit{v} is calculated for the respective macroblock and the MV estimation from the previous step is subtracted from it, leading to the Motion Vector Differences (MVD). This is then the information that will be sent over the bitstream and then decoded.
The decoding process happens in reverse, meaning the MVD are recovered, the MVs are estimated, and \textit{v}, our target MVs, are calculated by adding the two.

Knowing that the encoding process can be quite computationally expensive, hardware encoding and decoding solutions that automatize this process and completely remove the CPU-load, have become quite prevalent. There are even GPU hardware based decoding acceleration to support this process to allow for a faster decoding process without extra hardware. 
From there we surmise that using a hardware accessory that directly outputs the motion vectors into memory, it would be \textit{quasi-free}, in terms of CPU and GPU processing power, to get the MV information. However, as a proof of concept, we perform our following experiments using non-hardware-specific video decoding, as the source of the MVs should have no impact on the results.

\subsection{Data Preprocessing}
\label{method-datapre}

\textbf{Face detector.} Following \cite{caldelli2021optical,li2020face}, we use a pretrained MTCNN~\cite{zhang2016joint} face detector to detect the face region using the default parameters. Should multiple faces be detected, only the biggest one is used. After detecting the four corner points, the dimension that is smallest is then padded with pixels to obtain a square bounding box. The face is resized to a $224\times 224$ resolution. Finally the resulting frames are normalized using standardization.

\textbf{Motion Vectors.} After processing the faces, we extract the MVs and IMs in the face region and stack them together. 
We then end up utilizing either a four-dimensional input (past-x, past-y, future-x, future-y) to indicate whether the MV references a past or future frame and their direction, or a six-dimensional input (future- and past-referencing IMs) if the IM is employed as an additional input feature. 
The MVs are also normalized by standardization. 

There is however a caveat, in that there exist blocks which have a zero-MV, and blocks for which there is no MV information, that are also zero. As these non MV-blocks are represented by the IM we can exclude them from the normalization by referencing non-zero IM blocks. The MV and IM blocks are then scaled up to our selected input resolution, by using bi-linear and nearest neighbour interpolation respectively.

\begin{figure}[t]
    \centering
    \includegraphics[width=.48\textwidth,page=3]{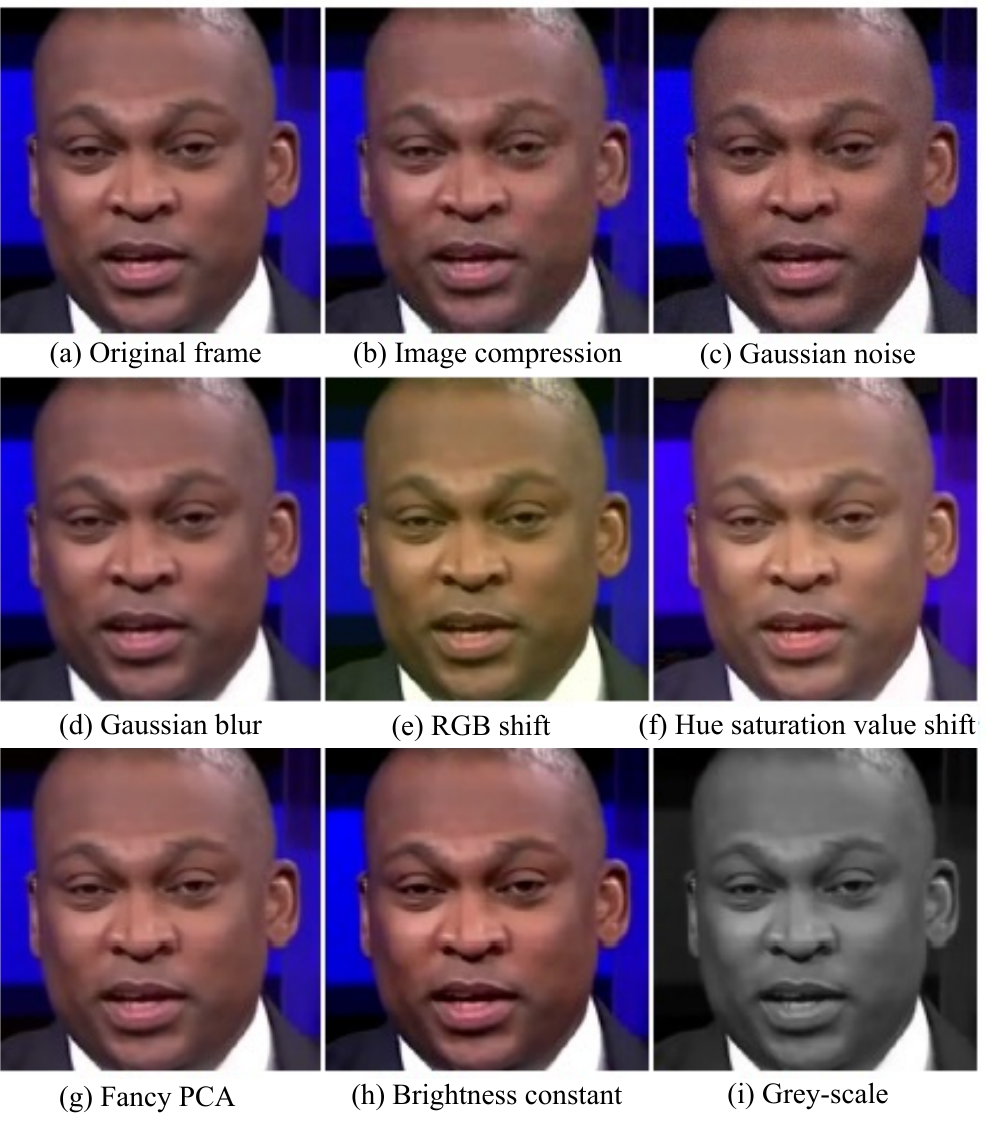}
    \caption{Examples of all the data augmentations performed on RGB data stemming from FaceForensics++.
    }
\label{fig:rgbaug}
\vspace{-12pt}
\end{figure}


\textbf{Data Augmentation.} We apply several augmentations, to both reduce overfitting and give us better predictions, inspired by the DFDC winner~\cite{kaggleDFDC}. We use the albumentation library~\cite{DBLP:journals/corr/abs-1809-06839} to implement the following augmentations, which are then applied with a certain probability each (see Figure~\ref{fig:rgbaug}): image compression, gaussian noise and blur, RGB and hue-saturation-value shifts, FancyPCA~\cite{NIPS2012_c399862d}, random Brightness Contrast, and grey-scale transformation.

\begin{figure}[t]
    \centering
    \includegraphics[width=.51\textwidth,page=8]{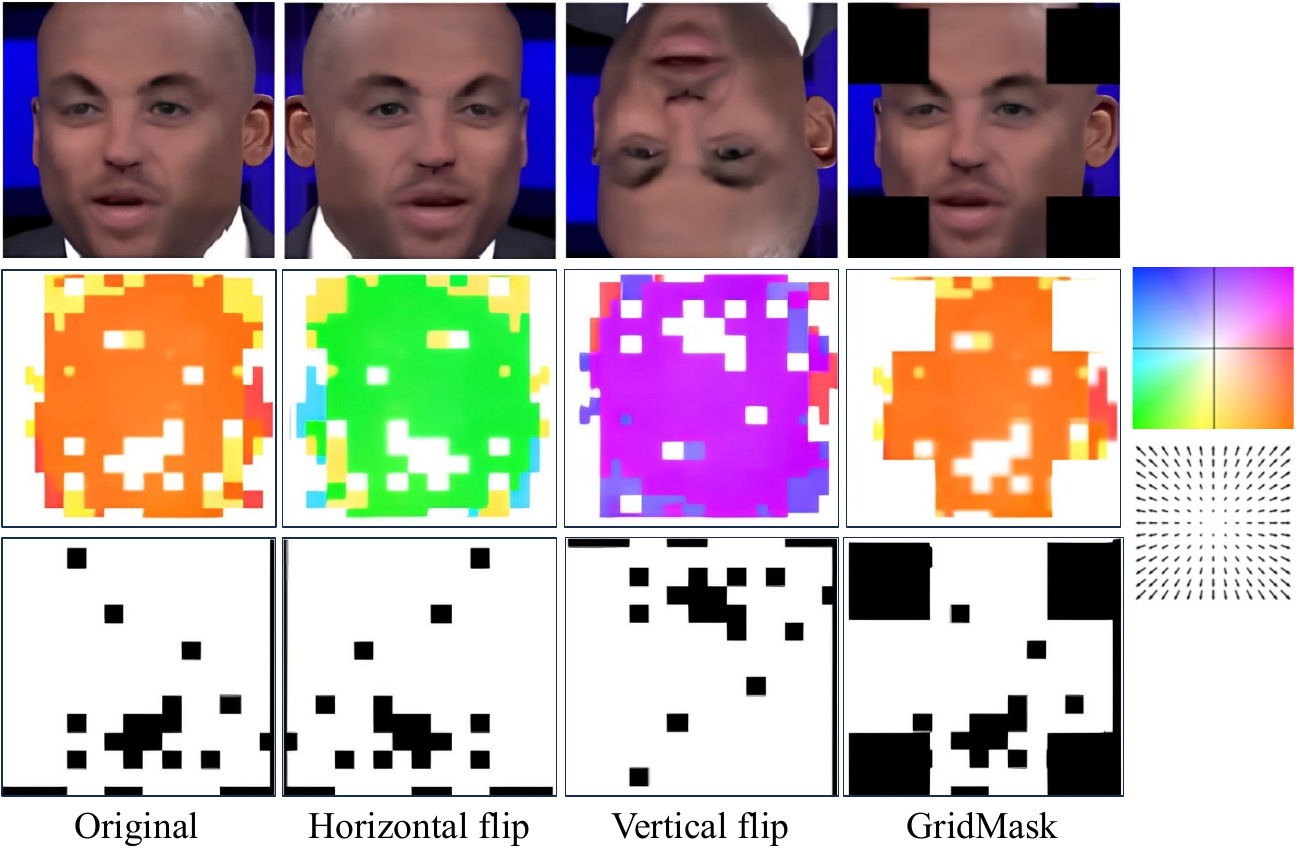}
    \caption{Examples of the data augmentations performed on motion vectors and information mask on the FaceForensics++ dataset. In these augmentations, the motion vector values have to be additionally mirrored on the flip axis in vector space, on top of flipping their positions. The colors represent the MVs direction and length in the vector field. (Best viewed on a screen when zoomed in)
    }
\label{fig:mvaug}
\vspace{-15pt}
\end{figure}

On top of these augmentations, we apply non-RGB transformations to the entirety of the image, which include horizontal and vertical flips, as well as the complete removal of certain regions of the input, replacing them by patches of zeroes as introduced in GridMask data augmentation~\cite{DBLP:journals/corr/abs-2001-04086}.
It is to note that flips and GridMask augmentations affect MVs differently, as their values need to be mirrored in vector space on top of flipping their position, which can be seen in Figure~\ref{fig:mvaug}.

\subsection{Classifier}
\label{method-classifier}

We choose MobileNetV3~\cite{howard2017mobilenets,howard2019searching} as our classifier's backbone. 
MobileNetV3 is designed using neural architecture search for low latency and accuracy, aligning with our goal of wanting to make our pipeline as efficient as possible. 
To use it for DeepFake detection, we made several modifications. 
Firstly, we replace the classifier output by a fully connected layer to output a scalar indicating the probability of the sample being fake. 
Secondly, we change the number of input channels of the first convolution, as we have varying channel sizes, depending on whether we use RGB, MVs, or MVs and IMs. 
The feature extraction part remains the same. 
Finally for our two-stream network, we combine the two different modalities of input, RGB and MVs by concatenating the last layers of their respective MobileNets, which consist of a single value. 
We then get our final prediction by averaging 100 randomly sampled frames of the video.

\textbf{Loss function.} DeepFake detection, as we define it, is a binary classification, and therefore we use the binary cross entropy loss.
\vspace{-10pt}
\begin{equation}
    \mathcal{L}=\hat y\log{y}+(1-\hat y)\log(1-y),
\end{equation}

where $\hat y$ and $y$ denotes the predicted and ground-truth label, respectively. 

\section{Experiments}
\label{experiments}


\subsection{Experimental setups}
\textbf{Datasets.} 
Following previous optical flow based DeepFake detection research~\cite{caldelli2021optical}, we report our findings on the FaceForensics++~\cite{rossler2019faceforensics++} dataset, which contains 1,000 YouTube videos that have been manipulated with different types of DeepFake manipulations, we use the hq version (C23). 
More specifically there are five different types: 
FaceShifter (FS)~\cite{li2020advancing}, FaceSwap (FSwap)~\cite{faceswap}, DeepFakes (DF)~\cite{originaldeepfakes}, Face2Face (F2F)~\cite{thies2016face2face}, and NeuralTexture (NT)~\cite{thies2019deferred}. The diversity of the data, and the pairing of fake and non-fake video of the same source, allow for practical initial explorations into generalizability of DeepFake detection methods, which is our goal.

\textbf{Baselines.} 
To fairly evaluate the performance of our motion vector based model, we compare it not only with pure RGB based models, but also with a state-of-the-art optical flow estimator, RAFT~\cite{teed2020raft}, based model, using the default parameters as described in the original paper. We create our own baselines, as we cannot compare to values from \cite{caldelli2021optical}, which does not report all quantitative values. Additionally, RAFT significantly outperforms TV-l1, the classical OF algorithm they use\footnote{\url{http://sintel.is.tue.mpg.de/results}}.
To avoid storing floating points, which would make for very large file sizes, we save the estimated optical flows as gray scale images using lossless jpeg compression as in \cite{simonyan2014two}.

\textbf{Implementation details.} 
We implement our model in PyTorch~\cite{NEURIPS2019_9015} and PyTorch Lightning~\cite{Falcon_PyTorch_Lightning_2019}.
The training and experiments are performed on two Nvidia GTX Titan X with 12 GB of memory and an Intel Xeon E5-2680 v3 @ 2.50GHz CPU. We use the Adam optimizer~\cite{kingma2017adam} and perform a balanced training, meaning depending on the dataset, the diverse DeepFake generated contents are sampled equally and sum up to the number of real samples.
The models are trained for eight epochs (until convergence) and the best performing checkpoint on validation loss is selected.



\begin{figure}[t]
    \centering
    \includegraphics[width=0.48\textwidth]{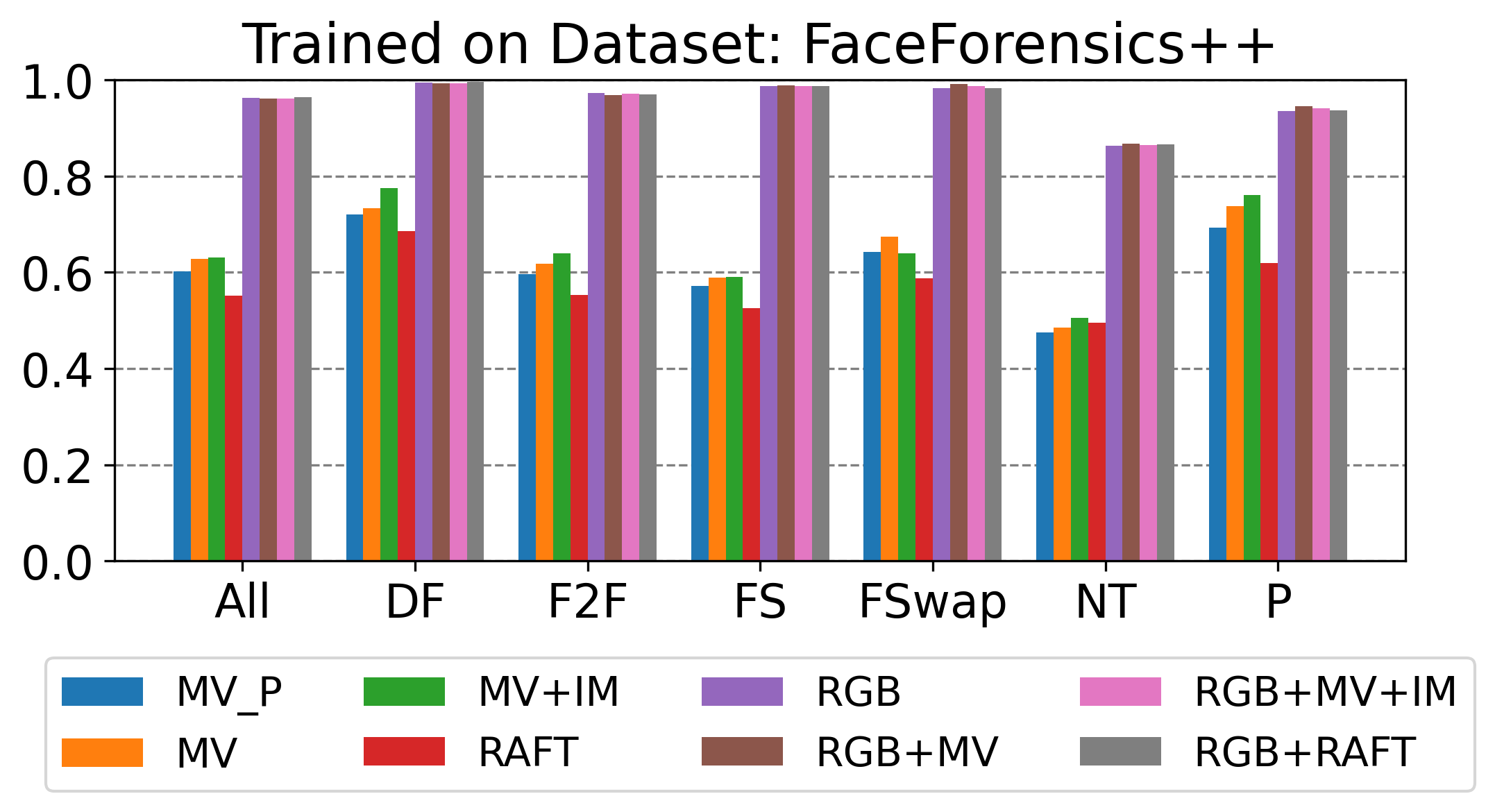}
    \caption{General evaluation of all our models on all FaceForensics++ DeepFake types. The values represent the accuracies on the respective testsets. We also examine $MV\_P$, which represents the use of only past frame referencing motion vectors, which would make it more closely matched to OF methods. $P$ represents the pristine (real) videos themselves. The values quickly overfit when combined with RGB, and adding auxiliary inputs only changes the accuracy by fractions of percentages.}

\label{fig:allacc}
\vspace{-15pt}
\end{figure}

\subsection{DeepFake detection accuracy}

To gain an understanding of their general accuracy in DeepFake detection, we evaluate the two-stream models for our temporally-aware models on the full dataset, and compare it to the RGB baseline (see Figure~\ref{fig:allacc}). The goal being to make the prediction as accurate as possible. 

Once we introduce RGB based models into the mix, we see an immediate saturation in accuracy for the models at around 96\%. Additionally, the RGB models once again quickly overfit on the dataset. As we see no changes in general accuracy whether the model uses RGB or a combination of RGB and motion information, we surmise that this is due to the poor quality of current DeepFake datasets available for research. Meaning the model gets as high of an accuracy it can get, already by just using RGB input.

To have a fairer evaluation of these combined models, we stress the importance of introducing better DeepFake detection datasets that include high quality real-world data.

\subsection{Cross-forgery generalization ability}

\begin{figure*}[t]
    \centering
    \includegraphics[width=1.\textwidth,page=1]{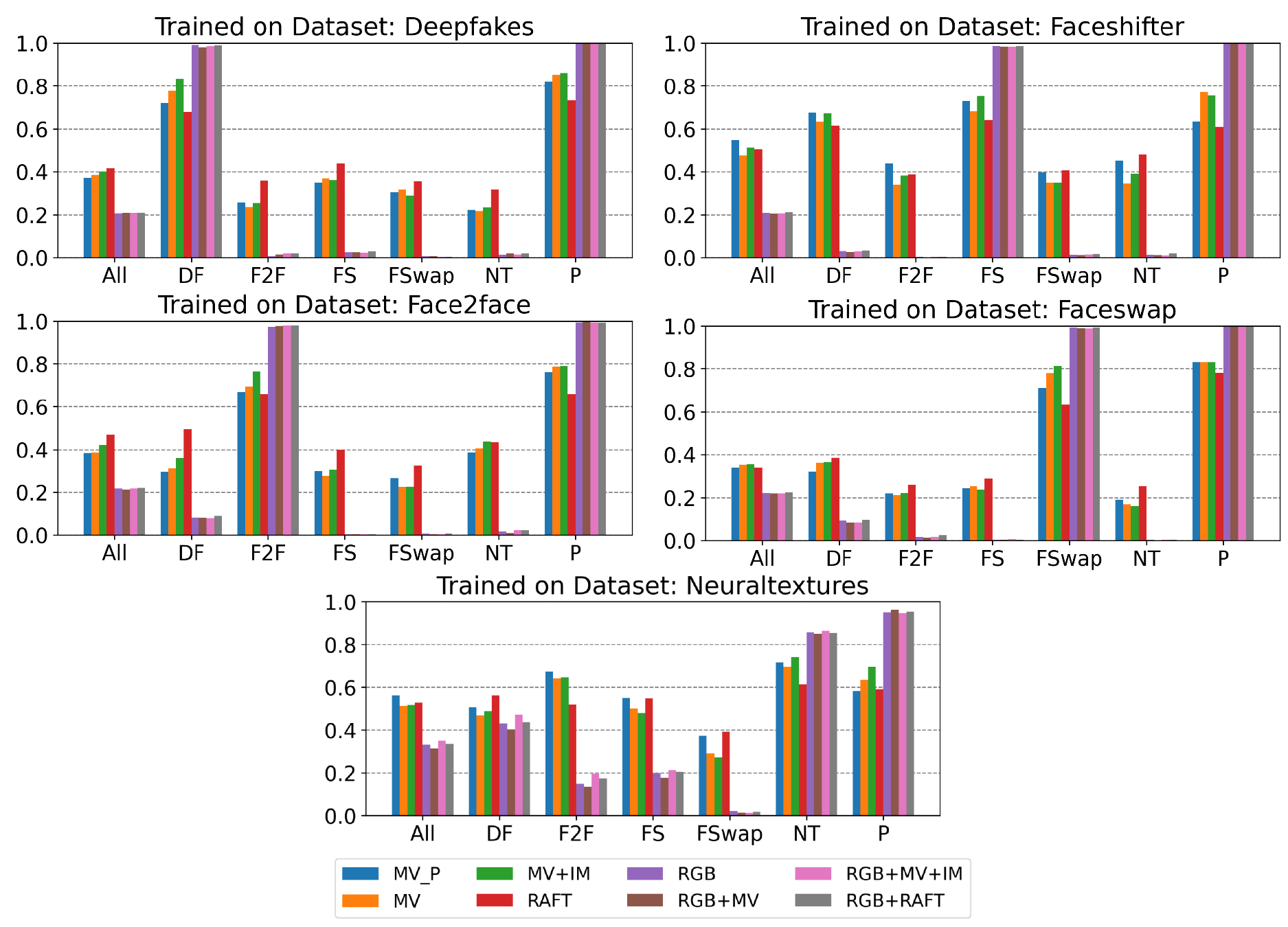}
    \caption{Cross-forgery evaluation of our models on datasets they were not trained for (out-of-distribution) in the FaceForensics++ dataset. The values represent the accuracies [0.0,1.0] of the models on the respective DeepFake or real (pristine) video test set.
    }
\label{fig:cross_eval}
\vspace{10pt}
\end{figure*}

A main focus in our experiments is to check whether we can maintain the generalizability that was initially shown when using optical flow models, over RGB models~\cite{caldelli2021optical}. To that end, we train our models on one type of DeepFake each, and evaluate them on all different DeepFake types in the dataset, as shown in Figure~\ref{fig:cross_eval}, Table~\ref{table:mvimcross} and Table~\ref{table:raftcross}.
Firstly, and as we elaborate in Section~\ref{onlytemp}, we can see MVs achieving higher accuracies on the specific datasets, indicating that they might add additional information that is not included in optical flows.
Secondly, for cross-forgery and on average we see equally strong generalization results for all temporal augmentations, with OF performing slightly better on some subsets, namely Face2Face and DeepFakes. Confirming previous research, we see purely RGB trained networks perform poorly on all generalization tasks.


Thereby, we confirm the cross-forgery detection ability of MV in DeepFake detection and perform a more in-depth comparison of their similarities in subsection~\ref{mvecapproxof}.

\subsection{Classification with only temporal data}
\label{onlytemp}

In this experiment we want to evaluate the performance of the classifier when fed with only temporal information.
Previous research~\cite{caldelli2021optical} reports average values of 82.99\% for this task, however due to the different model, optical flow estimator and test set, we cannot use these numbers as direct baseline.
Therefore we establish our own baseline by using the state-of-the-art RAFT optical flow as input and compare it with MVs and IMs concatenated with MVs (see Figure~\ref{fig:allacc} and Table~\ref{table:motion}).
We also evaluate the behavior of models relying solely on MVs stemming from the past (MV\_P), which would correspond more closely to the OF. 

These results show that using the MVs and IMs only, strongly outperforms the RAFT based model, even only using MV\_Ps. Meaning that, while maybe not all motion is accounted for by MVs as it is by the OF (see Section~\ref{mvecapproxof}), there is additional information in MVs and IMs that allows for a better classification of whether a video is a forgery or not.

\begin{table}[!h]
\centering

\resizebox{0.43\textwidth}{!}{%

\begin{tabular}{lccccc} \hline 
Method  & \multicolumn{1}{c}{DF} & \multicolumn{1}{c}{F2F} & \multicolumn{1}{c}{FS} & \multicolumn{1}{c}{FSwap} & \multicolumn{1}{c}{NT}  \\ \hline \hline
OF & 67.90 & 66.00 & 64.13 & 63.30 & 61.37 \\ \hline   
MV\_P  & 71.97 & 66.90 & 72.93 & 71.00 & 71.63 \\ \hline 
MV  & 77.60 & 69.50 & 68.20 & 77.93 & 69.53 \\ \hline 
MV+IM  & \textbf{83.53} & \textbf{76.50} & \textbf{75.30} & \textbf{81.17} & \textbf{74.23}  \\ \hline \\
\end{tabular}
}
\caption{Evaluation of motion vectors, information masks and optical flow classification accuracies (in \%) on the respective DeepFake types they were trained on.}
\label{table:motion}
\end{table}




\begin{table}[h]
    \centering
    \begin{tabular}{lrc}
    \hline
        Resolution & OF MFLOP & MV MFLOP  \\ \hline \hline
        480$\times$360 & 138.8 $\cdot$ 10$^{3}$ & 0.1  \\ \hline
        640$\times$480 & 249.3 $\cdot$ 10$^{3}$ & 0.2  \\ \hline
        1280$\times$720 & 783.5 $\cdot$ 10$^{3}$ & 0.6  \\ \hline
        1920$\times$1080 & 1.9 $\cdot$ 10$^{6}$ & 1.3 \\ \hline \\
    \end{tabular}
    \caption{Computational costs of creating optical flow based on RAFT compared with the costs of transforming the MVs into our input format, in MFLOPs.}
    \label{table:flop}
    \vspace{-10pt}
\end{table}

\begin{table*}[!h]
    \centering
    \resizebox{0.7\textwidth}{!}{
    \begin{tabular}{lcccccc}
    \hline
        MV+IM & DeepFakes & Face2Face & FaceShifter & FaceSwap & NeuralTextures & all  \\ \hline \hline
        DeepFakes & 83.53\% & 35.77\% & 67.23\% & 36.47\% & 48.73\% & \textbf{77.57\%}  \\ \hline
        Face2Face & 25.57\% & 76.50\% & 38.23\% & 22.20\% & 64.60\% & 63.93\%  \\ \hline
        FaceShifter & 36.10\% & 30.40\% & 75.30\% & 23.83\% & 47.87\% & 59.00\%  \\ \hline
        FaceSwap & 28.97\% & 22.57\% & 34.90\% & 81.17\% & 27.20\% & 64.00\%  \\ \hline
        NeuralTextures & 23.43\% & 43.63\% & 39.07\% & 16.23\% & \textbf{74.23\%} & 50.47\%  \\ \hline
        Pristine & \textbf{86.10\%} & \textbf{78.93\%} & \textbf{75.63\%} & \textbf{83.17\%} & 69.63\% & 76.03\%  \\ \hline
        all & 40.10\% & 42.10\% & 51.43\% & 35.53\% & 51.70\% & 63.03\% \\ \hline \\
    \end{tabular}
    }
    \caption{Cross-forgery accuracy evaluation of our MobileNet network with MV+IM input, trained on a specific DeepFake type (columns). Evaluated on different forgery types (rows).}
    \label{table:mvimcross}
    \vspace{10pt}
\end{table*}

\begin{table*}[!h]
    \centering
    \resizebox{0.7\textwidth}{!}{
    \begin{tabular}{lcccccc}
    \hline
        RAFT & DeepFakes & Face2Face & FaceShifter & FaceSwap & NeuralTextures & all  \\ \hline\hline
        DeepFakes & 67.90\% & 49.47\% & 61.53\% & 38.43\% & 56.20\% & \textbf{68.60\%}  \\ \hline
        Face2Face & 35.83\% & 66.00\% & 38.73\% & 25.97\% & 51.93\% & 55.23\%  \\ \hline
        FaceShifter & 43.63\% & 40.00\% & \textbf{64.13\%} & 28.97\% & 54.80\% & 52.57\%  \\ \hline
        FaceSwap & 35.43\% & 32.50\% & 40.63\% & 63.30\% & 39.13\% & 58.77\%  \\ \hline
        NeuralTextures & 31.73\% & 43.47\% & 47.87\% & 25.43\% & \textbf{61.37\%} & 49.53\%  \\ \hline
        Pristine & \textbf{73.33\%} & \textbf{66.07\%} & 61.00\% & \textbf{78.00\%} & 59.03\% & 61.90\%  \\ \hline
        all & 41.53\% & 46.87\% & 50.60\% & 34.00\% & 52.83\% & 55.13\% \\ \hline \\
    \end{tabular}
    }
    \caption{Cross-forgery accuracy evaluation of our MobileNet network with RAFT input as baseline, trained on a specific DeepFake type (columns). Evaluated on different forgery types (rows).}
    \label{table:raftcross}
    \vspace{10pt}
\end{table*}

\begin{figure*}[t]
\setlength{\belowcaptionskip}{-0.375cm}

    \centering
    \includegraphics[width=0.98\textwidth,page=6]{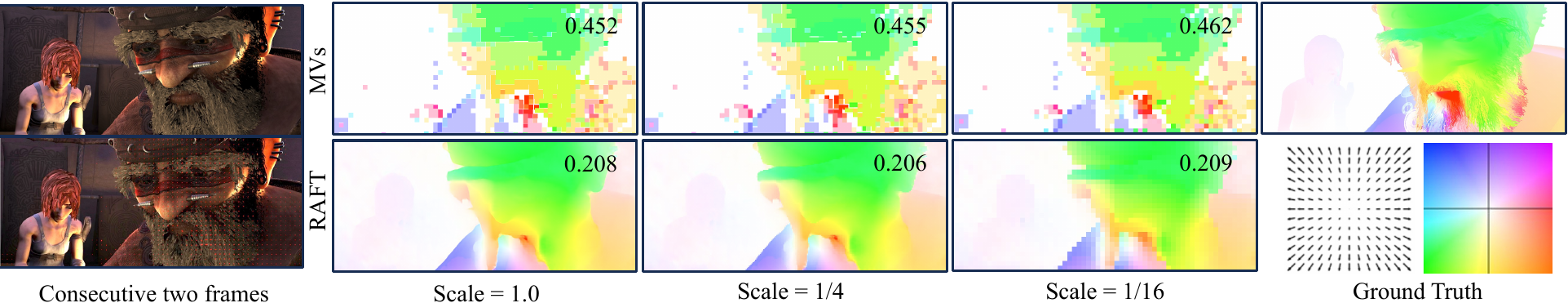}
    \caption{Comparison of MVs and optical flow, values indicates end-point-error shown. Image brightness is corrected for better visualization. (Best viewed on a screen when zoomed in)}

\label{fig:failure}
\vspace{-2pt}
\end{figure*}

\subsection{Run-time analysis}

We perform a general upper-bound analysis of the estimated operation cost of preparing motion vectors, as compared to obtaining the optical flow. We do this by firstly calculating the costs of transforming the motion vectors to fit our input type, assuming the encoded data is provided for free, and comparing it with RAFT FLOPs obtained by fvcore\footnote{\url{https://github.com/facebookresearch/fvcore}} given two input frames. Two frames as the OF algorithm needs to predict the motion between the two.
The different computational costs are visible in Table~\ref{table:flop}.

These values represent the difference between a single frame, therefore one can easily see that the processing time scales massively with a regular 30 fps stream from a webcam. 

\subsection{Using MVs as motion approximation}
\label{mvecapproxof}

Beyond the capabilities of MVs in DeepFake generalization, we also want to confirm the validity of MVs for motion approximation, which has already been used in prior research~\cite{yokoyama2009motion}. To evaluate their efficacy in this task, we assessed MVs on the Sintel optical flow benchmark~\cite{Butler:ECCV:2012}. Using the clean data split for simplicity, we generated MVs by compressing video frames into a video via FFmpeg~\cite{tomar2006converting} using the H.264 codec. Subsequently, in line with the optical flow evaluation, we computed the error of the MVs and tabulated the results in table~\ref{table::sintel}. 

We note that our computation only incorporated the P frames, as I frames lack motion information. We report end-point-error (EPE) in the table for different resolutions. We compare the different scales, as the MVs are per definition at a 16 times lower resolution. 
While the EPE results show significantly worse results for MV, they represent valid motion approximations by surpassing some of the current existing OF on OF benchmarks~\cite{Gehrig3dv2021}. It should also be taken into consideration that RAFT was trained on the Sintel dataset.


\begin{table}[h]
\centering
\setlength{\belowcaptionskip}{-0.4cm}


\begin{tabular}{ccc} \hline 
DownScale & Method  & \multicolumn{1}{l}{EPE $\downarrow$}    \\ \hline
\multirow{2}{*}{1/1} & RAFT & \textbf{0.603}    \\ \cline{2-3}
& MVs  & 2.193  \\ \hline 
\multirow{2}{*}{1/4} & RAFT  & \textbf{0.611}    \\ \cline{2-3}
& MVs  & 2.364 \\ \hline 
\multirow{2}{*}{1/16} & RAFT & \textbf{0.652}   \\ \cline{2-3}
& MVs  & 2.527  \\ \hline
\\
\end{tabular}
\caption{Evaluation of MVs as motion approximation. End-point-error (EPE) is reported. }
\label{table::sintel}
\vspace{-3pt}
\end{table}

\subsection{Ablation on data augmentation}

We run an ablation study on different types of data augmentation, and evaluate their contribution individually in Table~\ref{table:augm}. Training without augmentations led to quickly overfitting models, and adding augmentations generally gave us a performance and generalization boost.

\begin{table}[!ht]
    \centering
    \resizebox{0.35\textwidth}{!}{
    \begin{tabular}{ll}
    \hline
        Augmentations & Accuracy  \\ \hline
        None & 93.05\%  \\ \hline
        + Compression, Noise, Blur & 89.34\%  \\ \hline
        + Color changes &  93.62\%  \\ \hline
        + GridMask &  93.51\%  \\ \hline
        + Flips &  94.62\% \\ \hline \\
    \end{tabular}
    }
    \caption{Augmentations we use and their accuracy on the test set, when used with our two-stream model consisting of RGB+MV+IM streams.}
    \label{table:augm}
    \vspace{-3pt}
\end{table}

\section{Limitations and future work}

There are two noteworthy limitations in our method. First, we propose to replace optical flow with H.264 motion vectors for DeepFakes detection. Though it reduces computational costs by estimating the optical flow, we still have two efficiency bottlenecks in our pipeline: 
Firstly we still need to run face detection and classifier inference sequentially. Although we note that it would be possible to further exploit MVs for face detection and face tracking to boost prediction speed~\cite{yokoyama2009motion,wu2018compressed}. 
Secondly, as MVs have a 16 times lower resolution than the source video, it becomes significantly more challenging for the classifier when the potential subject of forgery does not occupy a larger part of the frame, or the frame has a low resolution to begin with. 
A solution we see, would be a compromise: Using MVs as priors to guide the estimation of optical flow as in \cite{wu2018compressed}. Since the coarse motion information would is provided by MVs, the optical flow estimation network can have a lighter design. 

%
While we show these capabilities for the most widely adapted H.264 codec, they are just as well portable to the newer codecs, such as H.265, which uses the Coding Tree Unit (CTU) structure instead of macroblocks, but still retains motion vectors.

\section{Conclusion}
\label{discussion}



In this work we have shown that the information available through the H.264 encoding-decoding process, can directly be used to augment existing RGB-only DeepFake detection pipelines, by including temporal information and artifacts. This allows for better generalization capabilities, comparable with optical flow based methods. Furthermore the temporal information is made available at a vastly reduced cost and can remove the need of running an optical flow network sequentially, before being able to run the rest of the detection pipeline. 
Finally, by leveraging hardware-based solutions for the decoding process, this auxiliary information becomes quasi-free. This, in turn, would enable real-time temporal anomaly detection even on consumer-grade hardware, highlighting its potential value in various applications such as video calls and streaming settings.

\textbf{Acknowledgement.} This work was supported in part
by the Swiss National Science Foundation via the Sinergia
grant CRSII5-180359.

\small
\printbibliography


\begin{biography}
\textbf{Peter Grönquist} is a Machine Learning Engineer at EPFL's Image and Visual Representation Lab. Starting his research at ETHZ in 2016, he specialized in computer vision at Huawei Research Zürich. At EPFL, he delves into image and video generation and detection. Besides academia, he consults with industry on DeepFake detection and promotes tech awareness and accessibility.
\\

\textbf{Yufan Ren} is a direct Ph.D. student (2020-) at EPFL's Image and Visual Representation Lab. He focuses on 3D vision and Neural Rendering and graduated from Zhejiang University, receiving the Chu Kochen Award. 
\\

\textbf{Qingyi He} is a MSc student in Computer Science at EPFL and is performing research in DeepFake analysis at the IVRL. She received her BSc in Computer Science from Zhejiang university.
\\

\textbf{Alessio Verardo} is a MSc student in Data Science at EPFL and has performed research on video DeepFake detection within IVRL. He received his BSc in communication systems at EPFL in 2021.
\\

\textbf{Sabine Süsstrunk} has directed EPFL's Image and Visual Representation Lab since 1999 and was the inaugural Director of the Digital Humanities Institute from 2015-2020. Specializing in computational photography, machine learning, and image quality. She's chaired numerous international conferences and is the President of the Swiss Science Council SSC, with roles in EPFL-WISH, SRG SSR, Largo Films and is a Fellow of IEEE and IS\&T.

\end{biography}

\end{document}